\documentclass{article}

\usepackage{microtype}
\usepackage{graphicx}
\usepackage{subcaption}
\usepackage{booktabs} %

\usepackage{hyperref}

\usepackage[preprint]{icml2026}

\usepackage{amsmath}
\usepackage{amssymb}
\usepackage{amsfonts}

\usepackage{amsmath,amsfonts,bm}

\def\eqref#1{equation~\ref{#1}}

\def\1{\bm{1}}

\DeclareMathAlphabet{\mathsfit}{\encodingdefault}{\sfdefault}{m}{sl}
\SetMathAlphabet{\mathsfit}{bold}{\encodingdefault}{\sfdefault}{bx}{n}

\usepackage[utf8]{inputenc} %
\usepackage[T1]{fontenc}    %
\usepackage{url}            %
\usepackage{nicefrac}       %
\usepackage{xcolor}         %
\usepackage{multirow}
\usepackage{wrapfig}
\usepackage{enumitem}
\usepackage{caption}

\newcommand{\hl}[1]{\textcolor{black}{#1}}
\newcommand{\ols}[1]{\mskip.5\thinmuskip\overline{\mskip-.5\thinmuskip {#1} \mskip-.5\thinmuskip}\mskip.5\thinmuskip} %

\icmltitlerunning{Probing Neural Topology of Large Language Models}

\begin{document}

\twocolumn[
\icmltitle{Probing Neural Topology of Large Language Models}

\begin{icmlauthorlist}
\icmlauthor{Yu Zheng}{mit}
\icmlauthor{Yuan Yuan}{nyu}
\icmlauthor{Yue Zhuo}{mit}
\icmlauthor{Yong Li}{thu}
\icmlauthor{Gabriel Kreiman}{havard}
\icmlauthor{Tomaso Poggio}{mit}
\icmlauthor{Paolo Santi}{mit}
\end{icmlauthorlist}

\icmlaffiliation{mit}{Massachusetts Institute of Technology, Cambridge MA, USA}
\icmlaffiliation{nyu}{New York University, New York City NY, USA}
\icmlaffiliation{thu}{Tsinghua University, Beijing, China}
\icmlaffiliation{havard}{Harvard Medical School, Boston MA, USA}

\icmlcorrespondingauthor{Yu Zheng}{yu\_zheng@mit.edu}

\icmlkeywords{Large Language Models, Interpretability, Neural Topology, Probing}

\vskip 0.3in
]

\printAffiliationsAndNotice{}  %

\begin{abstract}
Probing large language models (LLMs) has yielded valuable insights into their internal mechanisms by linking neural activations to interpretable semantics.
However, the complex mechanisms that link neuron's functional co-activation with the emergent model capabilities remains largely unknown, hindering a deeper understanding and safer development of LLMs.
In this work, we introduce graph probing, a method for uncovering the functional connectivity of LLM neurons and relating it to language generation performance.
By probing models across diverse LLM families and scales, we discover a universal predictability of language generation and understanding performance using only neural topology, which persists even when retaining just 1\% of neuron connections. 
\hl{Strikingly, probing on topology outperforms probing on activation by up to 130.4\% and 67.7\% on perplexity and space/time semantic regression respectively, suggesting that neural topology contains orders of richer information of LLM performance than neural activation, which can be easily extracted with simple linear or MLP probes.}
To explain the dependence between neural topology and language performance, we identify default networks and hub neurons in LLMs and provide causal evidence by interventional experiments on multiple benchmarks, showing that LLMs actually exploit these topological information.
Further analyses suggest that graph probing can be effectively leveraged to improve the efficiency and reliability of LLMs through proof-of-concept applications in model pruning and hallucination detection.
Codes and data for the graph probing toolbox are available at \url{https://github.com/DavyMorgan/llm-graph-probing}.
\end{abstract}

\section{Introduction}\label{sec::introduction}

Large language models (LLMs) exhibit remarkable generative capabilities~\citep{wei2022emergent,touvron2023llama,glm2024chatglm,team2024gemma,guo2025deepseek}, yet our understanding of how they succeed and what they have learned remains limited~\citep{sharkey2025open}. 
\hl{\textit{Probing}, which extract interpretable features from neural activations~\citep{alain2017understanding,rogers-etal-2020-primer,hewitt-liang-2019-designing,voita-titov-2020-information,pimentel-etal-2020-information}, has emerged as a powerful approach for reverse-engineering LLMs~\citep{belinkov2022probing, gurnee2024language}. }
For instance, Gurnee \textit{et al.}\citep{gurnee2024language} showed that LLMs encode a compact world model of space and time using linear regression probes. 
Unsupervised probing, such as sparse auto-encoders~\citep{engels2024not, gao2024scaling,rajamanoharan2024improving,lieberum2024gemma,mudide2024efficient} and cross-layer transcoders~\citep{dunefsky2024transcoders,lindsey2025biology}, have further revealed dictionaries of interpretable, mono-semantic concepts~\citep{huben24sparse} and even causal circuits~\citep{marks2024sparse}, corresponding to directions in neural latent space. 
While these advances shed light on the semantics of neural activations~\citep{sharkey2025open}, much less is known about how neurons are functionally connected, \textit{i.e.} the neural topology, which is believed to play an essential role in the emergence of intelligence~\citep{rathitopolm, bassett2017network}.

Recent studies have drawn compelling parallels between neurons in LLMs and those in the human brain~\citep{toneva2019interpreting,schrimpf2021neural,caucheteux2023evidence, kumar2024shared, rathitopolm, mischler2024contextual, tuckute2024driving, gahot2024fmri,sun2024brain,liu2025brain}, revealing shared properties such as spatial-functional organization~\citep{kumar2024shared,rathitopolm} and left lateralization~\citep{gahot2024fmri}. 
Neural activations at internal layers of LLMs have also been shown to reliably predict human brain responses given the same linguistic stimuli~\citep{schrimpf2021neural,tuckute2024driving,luo2022mapping}.
However, these efforts primarily focus on static neural activations of LLMs, while overlooking the key aspect of temporal and functional neural topology that has been studied in neuroscience for decades~\citep{bassett2006small, bassett2017network, fotiadis2024structure}. 
Moreover, although analogies between LLMs and human brains are insightful~\citep{toneva2019interpreting,goldstein2022shared}, few works explicitly connect these findings to the language generation performance, which is one of the primary indicators of an LLM's intelligence.

\hl{In this work, we introduce \textit{graph probing}, a novel approach for investigating the functional connectivity of neurons in LLMs and its relationship to language generation and understanding performance.}
By analyzing neural activity time series as LLMs process text token by token, we compute temporal and functional correlations between neurons to construct dynamic neural graphs.
\hl{Using this large-scale dataset of text-induced neural topology, we train probes to predict LLMs' accuracy in auto-regressively generating the corresponding text, as well as how LLMs represent fundamental concepts like space and time.}
\hl{In essence, graph probing connects the micro-level topology of how neurons are connected given a token sequence, to the macro-level performance of how well LLMs understand and predict these tokens, offering a new lens to study the emergent capabilities of LLMs.}
Our method is summarized in Figure \ref{fig::method} and described in Section \ref{sec::method}.

We then apply our graph probing framework to comprehensively analyze the neural topology of LLMs through extensive experiments.
\hl{First, we demonstrate that language understanding and generation performance can be reliably predicted using only the neural connectivity graph.} 
This predictability holds universally across LLM families and scales, outperforming activation-based probing approaches by up to 130.4\% even when preserving only 1\% of neuron connections, with empirical results spanning GPT~\citep{radford2019language}, Pythia~\citep{biderman2023pythia}, and Qwen~\citep{yang2024qwen2}, ranging from millions to billions of parameters (Section \ref{sec::results}).
Next, we show causal evidence through interventional experiments and analysis on the MMLU~\citep{mmlu,wang2024mmlu} benchmark, discovering stable default neural topology and hub neurons in LLMs, and more importantly, validating that LLMs actually utilize their internal topological information when generating text (Section \ref{sec::analysis}).
Finally, we offer two proof-of-concept applications of graph probing, showcasing its potential in model pruning and hallucination detection (Section \ref{sec::application}).
While not without limitations, we expect graph probing to provide valuable insights into the inner workings of LLMs and to guide their future development in an interpretable and reliable way.

\section{Graph Probing}\label{sec::method}

\begin{figure*}[t]
  \centering
  \includegraphics[width=0.85\linewidth]{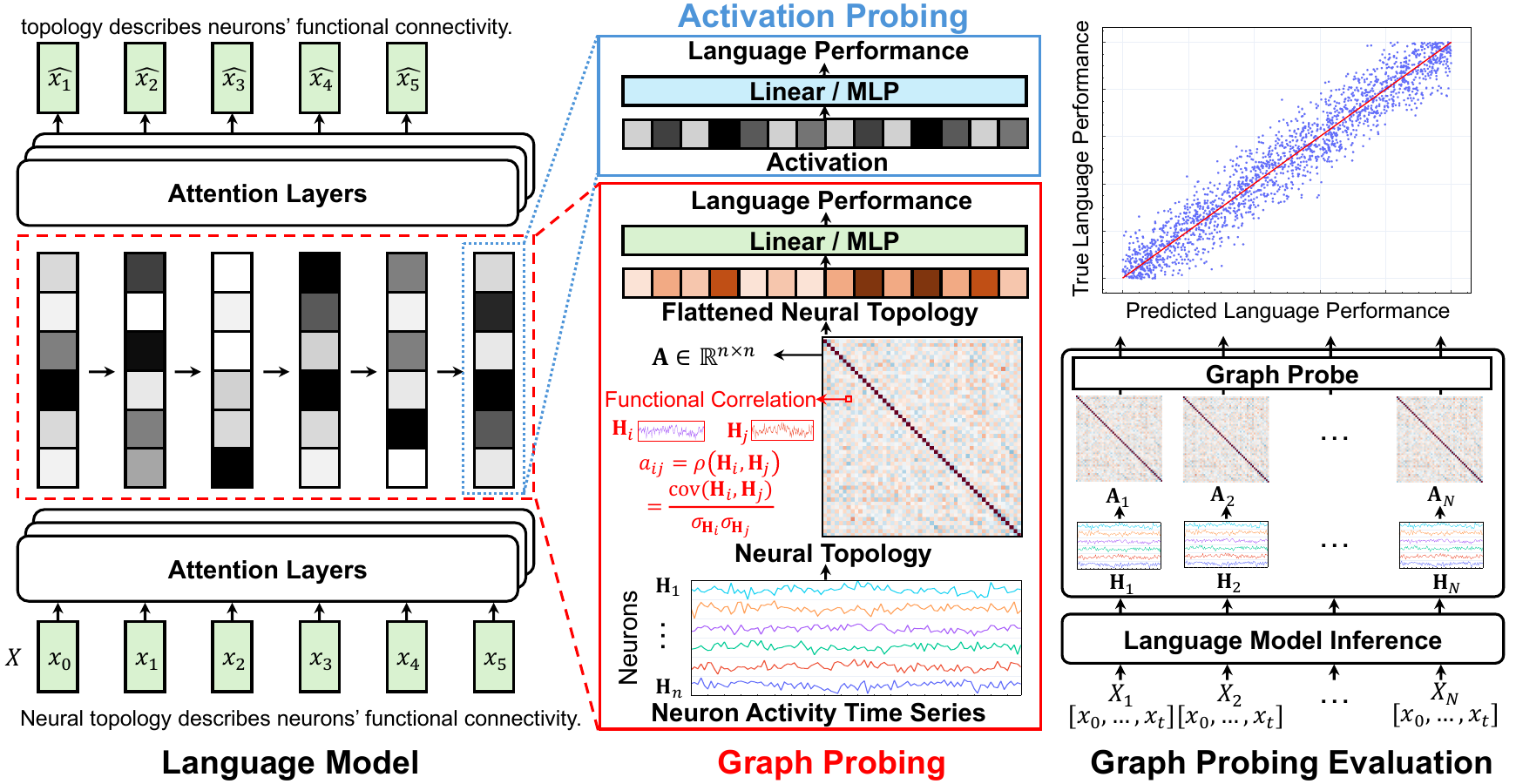}
  \vspace{-5px}
  \caption{Overview of graph probing. We extract the neuron activity time series in an LLM as it processes text token by token. We then compute temporal and functional correlations between neural activations to obtain topological connectivity graphs of neurons. Unlike existing probing methods that take neural activation as input, we train linear or MLP probes on flattened neural topology to predict the language generation performance for the input token sequence.}\label{fig::method}
  \vspace{-10px}
\end{figure*}

\paragraph{Neural Topology.}

To construct neural graphs from LLMs, we draw inspiration from neuroscience where functional brain networks are derived from temporal correlations in fMRI or EEG activation signals~\citep{bassett2017network, vertes2012simple, bullmore2011brain}, as shown in Figure~\ref{fig::method}.  
Given an LLM composed of stacked attention layers, the neural topology is constructed as follows:
\begin{align}
&\mathbf{H} = [\mathbf{h_0}, \mathbf{h_1}, \ldots, \mathbf{h_t}] \in \mathbb{R}^{n \times t},\label{eq::1}\\
&\mathbf{A} = \bigl(a_{ij}\bigr) \in \mathbb{R}^{n \times n}, \label{eq::2}\\
& a_{ij} = \rho(\mathbf{H}_{i,:}, \mathbf{H}_{j,:})\\
&=\frac{\sum_{k=0}^{t}{(\mathbf{H}_{i,k}-\ols{\mathbf{H}_{i,:}})(\mathbf{H}_{j,k}-\ols{\mathbf{H}_{j,:}})}}{\sqrt{\sum_{k=0}^{t}{(\mathbf{H}_{i,k}-\ols{\mathbf{H}_{i,:}})^2}}\sqrt{\sum_{k=0}^{t}{(\mathbf{H}_{j,k}-\ols{\mathbf{H}_{j,:}})^2}}},\label{eq::3}
\end{align}
where neurons at each layer produce a time series of hidden states $\mathbf{H}$ as the model processes a token sequence $X=[x_0,x_1,\ldots,x_t]$ \hl{($n$ and $t$ represent the number of neurons and tokens)}, and the temporal co-activation patterns among neurons define their \textit{functional connectivity}.
We capture this through a complete $n \times n$ weighted connectivity matrix $\mathbf{A}$, where each node corresponds to a neuron and each edge weight $a_{ij}$ represents the Pearson correlation coefficient between the activation time series of a pair of neurons $i$ and $j$~\citep{bassett2017network, fotiadis2024structure}.

\paragraph{Probing on Neural Topology.} 
We propose \textit{graph probing} to study the dependence between LLM performance and neural topology.
Specifically, we adopt simple linear or multi-layer perceptrons (MLP) probes~\citep{rumelhart1986learning} that take neural topology as input to predict its corresponding language understanding performance, as illustrated in Figure~\ref{fig::method}.
Given a connectivity matrix $\mathbf{A}$ induced by feeding a tokenized sequence $X$ to an LLM, where each element $a_{ij}$ denotes the functional connectivity (Pearson correlation coefficient) between neurons $i$ and $j$, our probe produces the graph representation $\mathbf{z}$ as follows:
\begin{alignat}{3}
    &\textbf{Linear}: ~ &\hat{p} = &\mathbf{W}_1 \cdot \mathtt{Flatten}(\mathbf{A})^T,\label{eq::linear_graph_probe}\\
    &\textbf{MLP}:~ &\hat{p} = &\mathbf{W}_3 \cdot \mathtt{ReLU}(\mathbf{W}_2 \cdot \mathtt{Flatten}(\mathbf{A})^T),\label{eq::mlp_graph_probe}
\end{alignat}
where $\mathtt{Flatten}(\mathbf{A})\in \mathbb{R}^{(n\times n)}$ is the flattened topology matrix, $\mathbf{W_1} \in \mathbb{R}^{1\times (n\times n)}, \mathbf{W_2} \in \mathbb{R}^{d\times (n\times n)}, \mathbf{W_3} \in \mathbb{R}^{1\times d}$ are weight matrices ($d$ as a hyper-parameter), $\hat{y}$ is the prediction of language performance.
In essence, the core and only difference of our method lies at the input, where we probe on neural topology instead of neural activation by existing approaches~\citep{belinkov2022probing,gurnee2023finding}.
We will later show that topology contains orders of richer information than activation regarding LLMs' intelligence.

\paragraph{Probing Target.}
\hl{By its definition, graph probing can be utilized to predict almost any fact of interest, and in this work we test it on various performance-related metrics, including perplexity, hallucination, and functional specialization, as well as fundamental semantics like space and time.}
Here we introduce the case of perplexity (PPL) as it is a fundamental metric directly reflecting LLMs' auto-regressive language generation performance~\citep{bengio2003neural}, while details of other cases are provided in later sections.
Specifically, given a token sequence, the perplexity is commonly calculated as the exponentiated average negative log-likelihood over the token sequence:
\begin{equation}
\mathtt{PPL}(X) = \exp\left\{ -\frac{1}{t} \sum_{i=1}^{t} \log p_\theta(x_i \mid x_{<i}) \right\}.
\end{equation}
The neural topology is dynamically induced by the specific token sequence, and our goal is to investigate whether the text-responsive neural topology is linked to how well the model predicts the text.
Towards this end, we train the graph probe to minimize the mean squared error (MSE) between predicted and true perplexities over a dataset of tokenized sequences $\mathbf{X}=\{X_1,\ldots,X_N\}$:
\begin{equation}
    \mathcal{L}(\mathbf{X}) = \frac{1}{N} \sum_{i=1}^N \left( \hat{p}_i - \mathtt{PPL}(X_i) \right)^2,\label{eq::mse}
\end{equation}
where the prediction $\hat{p}_i$ is calculated via the graph probe by Equation (\ref{eq::linear_graph_probe}) or (\ref{eq::mlp_graph_probe}).
Details of hyper-parameters and computational configurations are provided in Appendix \ref{app::method}.

\section{Results}\label{sec::results}

\noindent\textbf{LLMs.}
In our experiments, we train graph probes on neural topology derived from three families of LLMs, each spanning across different sizes.
Specifically, we evaluate GPT2~\citep{radford2019language} (GPT2, GPT2-large), Pythia~\citep{biderman2023pythia} (160M, 1.4B, 2.8B), and Qwen2.5~\citep{yang2024qwen2} (0.5B, 3B, 7B, 14B).
Details of the experimented LLMs are provided in Appendix \ref{app::llm}.

\noindent\textbf{Datasets.}
To enable our study, we construct neural topology using the OpenWebText dataset~\citep{Gokaslan2019OpenWeb}.
To ensure consistent temporal resolution, we control the length of neural activity time series to fall between 256 and 1024 tokens by merging consecutive sentences as needed. 
For each token sequence, we perform LLM inference to compute its perplexity and simultaneously extract hidden state time series to generate the corresponding neural topology.
For each model, we construct a probing dataset comprising about 10,000 graph–perplexity pairs.
Further details on dataset construction are provided in Appendix~\ref{app::dataset}.

\noindent\textbf{Evaluation.}
We split the dataset into training and test sets using an 8:2 ratio.
Having learned graph probes on the training set, we evaluate their out-of-sample prediction performance on the test set, which reveals the extent to which micro-level neural topology is predictive of macro-level language generation ability.
To quantify the effectiveness of graph probing, we report standard regression metrics on our test data, including mean squared error (MSE), mean absolute error (MAE), coefficient of determination ($R^2$), Pearson correlation ($\rho_p$), and Spearman rank correlation ($\rho_s$).
We compare with activation based baselines using both linear and MLP probes~\citep{alain2017understanding,belinkov2022probing,gurnee2024language,tuckute2024driving}.

\begin{figure*}[t]
  \centering
  \includegraphics[width=0.8\linewidth]{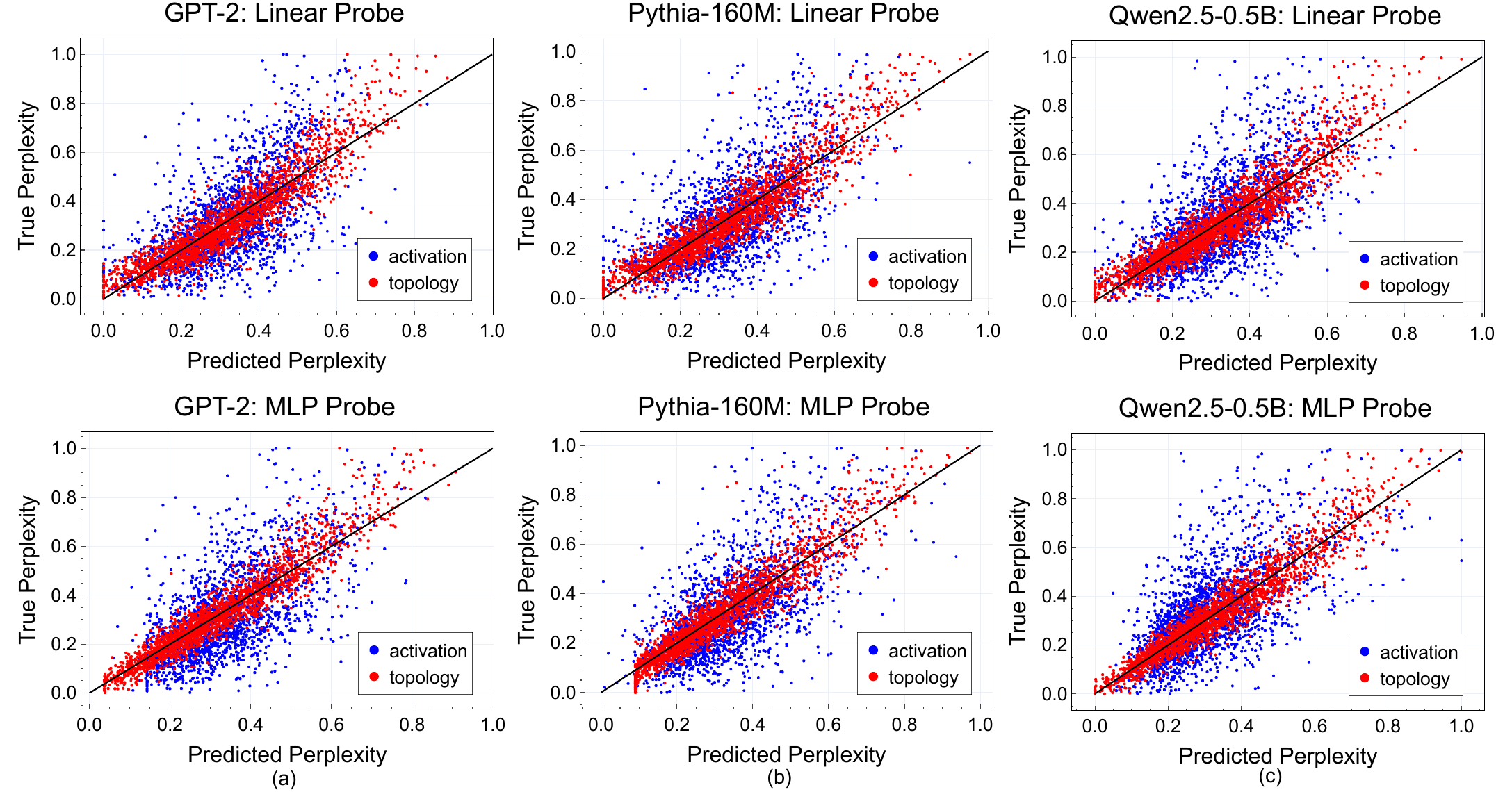}
  \caption{Out-of-sample performance of linear and MLP probing on the test set for (a) GPT-2 (b) Pythia-160M (c) Qwen2.5-0.5B. We compare activation-based probing and our topology-based probing. The correlation between the perplexity predicted by probing and the ground-truth perplexity reflects how well LLM performance can be inferred from neural activation or topology.}\label{fig::regression}
\end{figure*}

\begin{table*}[t]
\centering
\caption{Performance of different probing methods ($^*$ indicates $p$-value < 0.05).}
\label{tab:probe_performance}
\begin{tabular}{lllllll}
\toprule
\textbf{LLM} & \textbf{Probe} & \textbf{MSE $\downarrow$} & \textbf{MAE $\downarrow$} & \textbf{$\mathbf{R^2}$ $\uparrow$} & $\rho_p$ $\uparrow$ & $\rho_s$ $\uparrow$ \\
\midrule
\multirow{4}{*}{GPT-2} & Activation (Linear) & 0.0199 & 0.1067 & 0.3987 & 0.6352 & 0.6497 \\
 & Activation (MLP) & 0.0201 & 0.1056 & 0.3930 & 0.6370 & 0.6426 \\
 & Graph (Linear) & \textbf{0.0049}$^*$ & \textbf{0.0526}$^*$ & \textbf{0.8517}$^*$ & \textbf{0.9229}$^*$ & \textbf{0.9320}$^*$ \\
 & Graph (MLP) & \textbf{0.0031}$^*$ & \textbf{0.0399}$^*$ & \textbf{0.9057}$^*$ & \textbf{0.9534}$^*$ & \textbf{0.9577}$^*$ \\
 \midrule
\multirow{4}{*}{Pythia-160M} & Activation (Linear) & 0.0199 & 0.1061 & 0.4151 & 0.6517 & 0.6504 \\
 & Activation (MLP) & 0.0194 & 0.1030 & 0.4304 & 0.6618 & 0.6519 \\
 & Graph (Linear) & \textbf{0.0047}$^*$ & \textbf{0.0518}$^*$ & \textbf{0.8607}$^*$ & \textbf{0.9278}$^*$ & \textbf{0.9364}$^*$ \\
 & Graph (MLP) & \textbf{0.0036}$^*$ & \textbf{0.0432}$^*$ & \textbf{0.8930}$^*$ & \textbf{0.9458}$^*$ & \textbf{0.9499}$^*$ \\
\midrule
\multirow{3}{*}{Qwen2.5-0.5B} & Activation (Linear) & 0.0190 & 0.1028 & 0.4225 & 0.6536 & 0.6632 \\
 & Activation (MLP) & 0.0196 & 0.1045 & 0.4044 & 0.6496 & 0.6457 \\
 & Graph (Linear) & \textbf{0.0045}$^*$ & \textbf{0.0496}$^*$ & \textbf{0.8640}$^*$ & \textbf{0.9296}$^*$ & \textbf{0.9422}$^*$ \\
 & Graph (MLP) & \textbf{0.0030}$^*$ & \textbf{0.0396}$^*$ & \textbf{0.9095}$^*$ & \textbf{0.9538}$^*$ & \textbf{0.9583}$^*$ \\
\bottomrule
\end{tabular}
\vspace{-10px}
\end{table*}

\begin{table*}[t]
\tiny
\centering
\caption{\hl{Probing results on Arts (time) and World Places (space) datasets. We compare Average Activation and our Graph Probing method ($^*$ indicates $p$-value $<$ 0.05).}}
\label{tab:probing_results}
\resizebox{\textwidth}{!}{%
\begin{tabular}{llcccccc}
\toprule
\multirow{2}{*}{\textbf{LLM}} & \multirow{2}{*}{\textbf{Probe}} & \multicolumn{3}{c}{\textbf{Arts (time)}} & \multicolumn{3}{c}{\textbf{World Places (space)}} \\
\cmidrule(lr){3-5} \cmidrule(lr){6-8}
 & & MSE & MAE & $R^2$ & MSE & MAE & $R^2$ \\
\midrule
\multirow{4}{*}{\textbf{GPT-2}} 
 & Activation (Linear) & 0.0406 & 0.1596 & 0.2945 & 0.0360 & 0.1957 & 0.5106 \\
 & Activation (MLP) & 0.0382 & 0.1511 & 0.3359 & 0.0311 & 0.1727 & 0.5733 \\
 & \textbf{Graph (Linear)} & \textbf{0.0326}$^*$ & \textbf{0.1371}$^*$ & \textbf{0.4341}$^*$ & \textbf{0.0317}$^*$ & \textbf{0.1765}$^*$ & \textbf{0.5689}$^*$ \\
 & \textbf{Graph (MLP)} & \textbf{0.0267}$^*$ & \textbf{0.1230}$^*$ & \textbf{0.5361}$^*$ & \textbf{0.0258}$^*$ & \textbf{0.1514}$^*$ & \textbf{0.6486}$^*$ \\
\midrule
\multirow{4}{*}{\textbf{Pythia-160M}} 
 & Activation (Linear) & 0.0399 & 0.1580 & 0.3071 & 0.0360 & 0.1968 & 0.5116 \\
 & Activation (MLP) & 0.0391 & 0.1560 & 0.3211 & 0.0306 & 0.1705 & 0.5812 \\
 & \textbf{Graph (Linear)} & \textbf{0.0282}$^*$ & \textbf{0.1275}$^*$ & \textbf{0.5100}$^*$ & \textbf{0.0267}$^*$ & \textbf{0.1583}$^*$ & \textbf{0.6335}$^*$ \\
 & \textbf{Graph (MLP)} & \textbf{0.0266}$^*$ & \textbf{0.1234}$^*$ & \textbf{0.5387}$^*$ & \textbf{0.0246}$^*$ & \textbf{0.1447}$^*$ & \textbf{0.6636}$^*$ \\
\midrule
\multirow{4}{*}{\textbf{Qwen2.5-0.5B}} 
 & Activation (Linear) & 0.0325 & 0.1406 & 0.4359 & 0.0296 & 0.1747 & 0.6002 \\
 & Activation (MLP) & 0.0314 & 0.1368 & 0.4555 & 0.0268 & 0.1574 & 0.6367 \\
 & \textbf{Graph (Linear)} & \textbf{0.0258}$^*$ & \textbf{0.1221}$^*$ & \textbf{0.5524}$^*$ & \textbf{0.0215}$^*$ & \textbf{0.1364}$^*$ & \textbf{0.7080}$^*$ \\
 & \textbf{Graph (MLP)} & \textbf{0.0268}$^*$ & \textbf{0.1254}$^*$ & \textbf{0.5341}$^*$ & \textbf{0.0223}$^*$ & \textbf{0.1398}$^*$ & \textbf{0.6953}$^*$ \\
\bottomrule
\end{tabular}%
}
\end{table*}

\begin{figure*}[t]
  \centering
  \includegraphics[width=\linewidth]{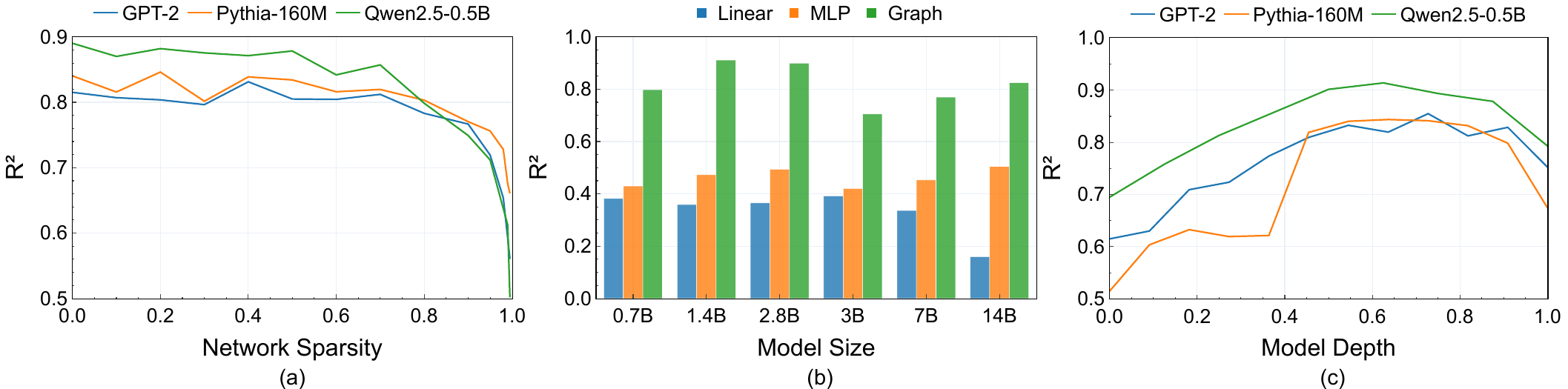}
  \caption{(a) Out-of-sample graph probing performance on neural topology of different sparsity levels. (b) Out-of-sample probing performance on LLMs of different sizes. (c) Out-of-sample performance of graph probing on different layers of LLMs.}\label{fig::property}
  \vspace{-10px}
\end{figure*}

We visualize the predicted and groundtruth perplexity in Figure~\ref{fig::regression} and summarize the results in Table \ref{tab:probe_performance}, where graph probing consistently outperforms activation probing across all three LLM families for both linear and MLP probes.
The improvements are strikingly substantial, with the maximum progress in $R^2$ exceeding 130.4\%.
For example, perplexity is barely predictable from neural activation, with $R^2$ less than 0.45 for all models, while that of graph probing all close to or even larger than 0.90.
The enormous gain of graph probing validates the hypothesis that neural topology contains much richer information of LLMs' language generation performance than neural activation, which can be easily extracted using simple linear or MLP probes.

\hl{
Besides predicting perplexity, we also extend our analysis to include a wider range of semantics, probing neural topology to predict time and space context of the input text.
We adopt the space and time datasets utilized by \cite{gurnee2024language}. 
We construct the neural topology for input texts regarding historical artworks and world places, and trained probes to predict: (1) time: the release year of historical artworks; (2) space: the longitude and latitude of world places. 
The results are summarized in Table \ref{tab:probing_results}, which serve two critical purposes.
Firstly, they validate findings from previous work~\cite{gurnee2024language} showing that LLMs possess internal representations of time and space. 
Secondly and crucially, they confirm that neural topology contains significantly richer information than neural activation for these semantic tasks. 
The substantial performance margin of as much as 67.77\% suggests that the structure of computation (how neurons connect) is more predictive than the magnitude of computation (activations) for extracting high-level concepts.
}

\noindent\textbf{Sparsity and scalability.}
Probing on complete graphs, \textit{i.e.}, dense $n \times n$ connectivity matrices that capture pairwise functional correlations between all neurons, can become computationally prohibitive as the LLM size increases, due to the quadratic number of edges that directly impacts the computational cost in both time and memory.
For instance, while complete graph probing is feasible for Pythia-160M with 768 neurons and 0.6M edges per layer, the number of edges in Qwen2.5-14B--comprising 5,120 neurons per layer--explodes to over 26M per graph.
\hl{To address this, on the one hand, we investigate whether probes can be applied to sparse graphs with weakly correlated edges pruned out by thresholding, which is commonly employed in human brain network construction~\citep{bassett2017network} and such sparsity is frequently observed in artificial neural networks~\citep{vig2020investigating,timkey-van-schijndel-2021-bark,dettmers2022gpt3}; on the other hand, we develop graph neural network probe~\citep{kipf2017gcn} that reduces the number of probe parameters from $\mathcal{O}(n^2)$ to $\mathcal{O}(n\cdot d)$ by weight sharing (details in Appendix \ref{app::gcn}).}
We train graph probes on neural topology with varying levels of sparsification (Figure~\ref{fig::property}(a)) for the perplexity task.
Surprisingly, the predictive performance remains remarkably stable even after removing up to 90\% of the edges, with minimal degradation.
Notably, even under extreme sparsity where only 1\% of the original edges are retained, the neural topology still enables effective prediction of perplexity, achieving above 0.6 $R^2$ that is still higher than activation probing.

The above experiments suggest that most of the predictive signal resides in a small subset of strong connections, making it possible to significantly reduce the number of edges while preserving nearly all critical topological information.
Leveraging this insight, we scale up graph probing to much larger models by operating on sparsified neural topology.
\hl{While our earlier results focused on models with fewer than 0.5B parameters, we now train probes on sparse graphs derived from LLMs with up to 14B parameters (GPT2-large, Pythia-\{1.4B, 2.8B\}, Qwen2.5-\{3B, 7B, 14B\}).}
As shown in Figure~\ref{fig::property}(b), graph probing continues to exhibit strong regression performance across all six large models, achieving a maximum $R^2$ of over 0.91, providing compelling evidence that the relationship between neural topology and language modeling performance is universal across model sizes.
Particularly, the gap between baselines and graph probing remains as substantial as 92.6\%, confirming the informativeness of neural topology.

\begin{figure*}[t]
  \centering
  \includegraphics[width=\linewidth]{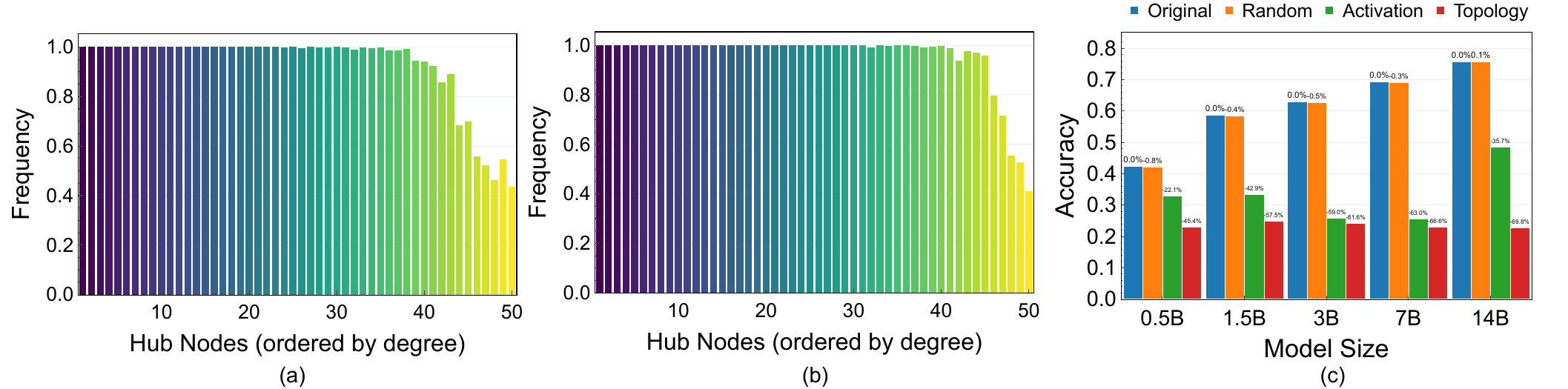}
  \caption{(a-b) Occurrence frequency of hub nodes in (a) Qwen2.5-0.5B and (b) Qwen2.5-1.5B on MMLU benchmark. \hl{(c) Accuracy on MMLU benchmark of Qwen2.5 models (0.5B, 1.5B, 3B, 7B, 14B) under different interventions of top 1\% neurons.}}\label{fig::mmlu}
  \vspace{-10px}
\end{figure*}

\noindent\textbf{Probing different layers.}
We further train probes on neural topology derived from different layers of GPT-2, Pythia-160M, and Qwen2.5-0.5B models.
Figure~\ref{fig::property}(c) reveals that while topologies from all layers are predictive of LLM performance in terms of perplexity, the middle layers are consistently the most informative across all three models, with $R^2$ values exceeding 0.8. 
This finding aligns with prior work identifying the middle layers of LLMs as the hub of semantic processing, whereas the initial and final layers are more specialized for \{de, re\}-tokenization~\citep{gurnee2024language, tuckute2024driving}.

\section{Topological analysis and causal intervention}\label{sec::analysis}

While our probing experiments discover the correlation between neural topology and language generation performance, the specific underlying topological structures remain unclear. 
In neuroscience, human brains exhibit a stable intrinsic network--a core connectivity pattern that persists to great extent across different tasks and stimuli~\citep{fotiadis2024structure,cole2014intrinsic}. 
Inspired by this, we investigated whether a similar default network of highly connected neurons exists within LLMs.
To test this hypothesis, we first calculated an average neural topology for the Qwen2.5-0.5B and Qwen2.5-1.5B models over the entire OpenWebText dataset. 
From this average, we identified the top 50 ``hub'' neurons (those with the highest degree) and then measured their occurrence frequency, \textit{i.e.}, how often these same neurons ranked as hubs in the topology of each individual data sample. 
The results, shown in Figure \ref{fig::mmlu}(a-b), are striking. 
A core set of nearly 40 neurons remain hubs in 100\% of the topologies across the entire dataset. 
This remarkable stability confirms the existence of a default network in LLMs, where a fixed set of hub neurons play dominant roles regardless of the input, suggesting a fundamental organizing principle of their internal structure.

To determine if LLMs rely on their neural topology, we conducted a series of interventional experiments. 
\hl{We disabled a selected 1\% of neurons in the middle layer of Qwen2.5 models (0.5B, 1.5B, 3B, 7B, 14B) by pinning their activations to zero at all tokens and measured the impact on the MMLU benchmark~\citep{mmlu}.}
We compared three distinct neuron selection strategies: random selection, selecting neurons with the highest activation, and selecting neurons with the highest topological degree.
As shown in the Figure \ref{fig::mmlu}(c), the results demonstrate a clear hierarchy of neuronal importance. While disabling random neurons caused a negligible accuracy drop (<1.0\%), targeting neurons by either activation or topology incurred a substantial performance collapse of at least 20\%. 
\hl{Critically, the topology-based intervention was the most detrimental. 
Disabling the top 1\% of hub neurons resulted in a catastrophic accuracy drop of 45.4\%, 57.5\%, 61.6\%, 66.6\%, and 69.8\% for the five tested models, consistently outperforming the activation-based strategy across all scales.
Particularly, for the largest 14B model, the relative performance drop against activation based intervention is over 95.5\%.
} 
These experiments provide causal evidence that LLMs actively utilize their underlying topological structure, particularly their hub neurons, for computation.

Our analysis also revealed a notable phenomenon of functional specialization~\citep{cole2014intrinsic}, where neural topology displays distinct patterns for different subjects. 
We found a striking example by identifying certain neuron that consistently specializes in STEM subjects, while remaining almost entirely dormant for social science subjects. 
This suggests a subject-based organization within LLMs, and the details are provided in Appendix \ref{app::functional_specialization}.

\section{Applications}\label{sec::application}

\begin{figure*}[t]
  \centering
  \includegraphics[width=\linewidth]{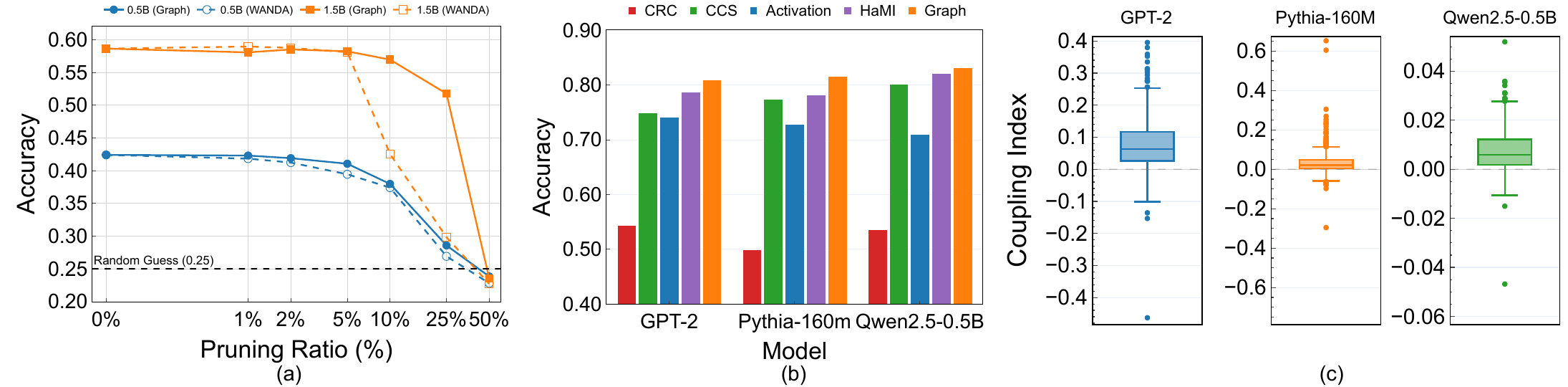}
  \caption{\hl{(a) Accuracy on MMLU benchmark under different levels of model pruning based on neural topology and activation (WANDA). (b) Accuracy of hallucination detection of different approaches on TruthfulQA dataset. (c) Coupling index of neural topology for hallucination on TruthfulQA datset.}}\label{fig::application}
  \vspace{-10px}
\end{figure*}

\noindent\textbf{Model pruning.}
\hl{Model pruning is a standard technique for optimizing LLMs in real-world deployments that prioritize low latency and computational efficiency~\citep{DBLP:conf/iclr/Sun0BK24,ma2023llm,gao2024disp,muralidharan2024compact,DBLP:conf/iclr/XiaGZ024,DBLP:conf/iclr/XuSCTZ0A0024}.}
The finding that high-degree neurons are functionally dominant (Section \ref{sec::analysis}) naturally suggests a pruning strategy of disabling low-degree neurons.
To test this, we prune these neurons by zeroing out their activations during inference. We apply this technique to a middle layer of Qwen2.5-0.5B and Qwen2.5-1.5B and evaluate the performance on the MMLU benchmark~\citep{mmlu}.
\hl{We compare our topology based pruning approach with a state-of-the-art LLM pruning method WANDA~\citep{DBLP:conf/iclr/Sun0BK24} which is based on neural activation.}
The results, presented in Figure \ref{fig::application}(a), confirm the models' robustness. 
Performance degradation is minimal even with substantial pruning; for the 1.5B model, accuracy drops by just 2.86\% when 10\% of neurons are pruned, and by 11.73\% when 25\% are removed. 
Strikingly, the models maintain above-random-guess accuracy until half of the low-degree neurons are pruned. 
\hl{In addition, our topology based pruning approach consistently outperforms WANDA, particularly in the 1.5B model where our method retains over 51\% accuracy compared to WANDA's 30\% accuracy when pruning 25\% of the neurons.}
These results highlight the potential for developing more sophisticated pruning methods based on neural topology.

\noindent\textbf{Hallucination detection.}
\hl{Building on our finding that subtle topological patterns in LLMs can predict text generation performance, we now investigate if these patterns can also detect hallucination, a critical challenge that has been extensively explored through various probing techniques~\citep{DBLP:conf/iclr/BurnsYKS23,du2024haloscope,hou2025probabilistic,DBLP:conf/acl/SuWAH00024,sriramanan2024llm,DBLP:conf/iclr/OrgadTGRSKB25,farquhar2024detecting,niu2025robust}.}
To create distinct \textit{genuine} and \textit{hallucinating} states, we construct inputs from the TruthfulQA dataset~\citep{lin2022truthfulqa} by concatenating each of its 817 question with its corresponding true and false answers, respectively, resulting in a dataset of 5918 samples (true/false classification). 
We then extract the neural topology from the LLM as it processes these inputs (Equations \ref{eq::1}-\ref{eq::3}). 
For this task, we adapt probing from regression to binary classification by modifying the MLP probe layer to have two output channels ($\mathbf{W}_1 \in \mathbb{R}^{2\times(n\times n)}$ and $\mathbf{W}_3 \in \mathbb{R}^{2\times d}$ in Equations (\ref{eq::linear_graph_probe}-\ref{eq::mlp_graph_probe})) and replacing the MSE loss with a cross-entropy loss:
\begin{alignat}{3}
    &\mathcal{L}(\mathbf{X}) = &\frac{1}{N} \sum_{i=1}^N \mathtt{CROSS\_ENTROPY}(\hat{y}_i, y_i),\label{eq::loss_hallucination}
\end{alignat}
where $\hat{y}$ and $y$ are the prediction and ground-truth for hallucination (0/1).
\hl{We compare against MLP probes trained on the neural activation at the last-token position.
We also include the state-of-the-art HaMI~\citep{niu2025robust} and CCS~\citep{DBLP:conf/iclr/BurnsYKS23} hallucination detection approach, as well as CCS's variant CRC~\citep{DBLP:conf/iclr/BurnsYKS23} for comparison.
}

We split the dataset into training and test sets using an 8:2 ratio to evaluate our hallucination detection framework. 
\hl{As shown in Figure \ref{fig::application}(b), probing neural topology substantially outperforms all baselines probing neural activation for all three tested models (GPT-2, Pythia-160M, and Qwen2.5-0.5B), with accuracy gains of up to 9.73\%.} 
This superior performance suggests that distinct topological patterns emerge when an LLM is generating a factual response versus hallucinating. 
To validate this hypothesis directly, we introduce a neural topology coupling index, 
\begin{align}
    &C_{XY} = \mathtt{AVG}(\{\rho(A_i,A_j) | A_i \in A_X, A_j \in A_Y\}), \\
    &C = C_{TT} + C_{HH} - 2C_{TH},
\end{align}
where the index $C$ measures the difference between intra-group similarity ($C_{TT}$, $C_{HH}$) and inter-group similarity ($C_{TH}$).
Here $A_T$ and $A_H$ represent the sets of neural topologies for truthful and hallucinated responses, respectively, and similarity is measured by the Pearson correlation ($\rho$) between each pair of the flattened adjacency matrices. 
We calculated this index for each question in the dataset, and the distribution in Figure \ref{fig::application}(c) shows that over 80\% of samples have a positive coupling index. 
This confirms that topologies are indeed more similar within the same state (truthful-to-truthful, hallucinated-to-hallucinated) than across different states (truthful-to-hallucinated), implying that neural topology serves as a promising and reliable signature for detecting LLM hallucinations and paving the way for future work on improving model reliability.

It is worthwhile to emphasize that our goal with such a simple topological probe is not to establish a new state-of-the-art against more nuanced approaches specifically design for either model pruning or hallucination detection, which are highly specialized domains with extensive literature~\citep{DBLP:conf/iclr/Sun0BK24,ma2023llm,gao2024disp,muralidharan2024compact,DBLP:conf/iclr/XiaGZ024,DBLP:conf/iclr/XuSCTZ0A0024,kuhnsemantic,niu2025robust,DBLP:conf/iclr/BurnsYKS23,du2024haloscope,farquhar2024detecting,sriramanan2024llm}.
Rather, these experiments serve as proof-of-concept demonstrations. 
Though preliminary in their current forms, these case studies showcase the versatile utility of graph probing and lay the groundwork for future work to translate these concepts into robust, system-level solutions.
We hope these results encourage the community to view graph probing as a valuable tool for optimizing and interpreting LLMs.

\section{Related Work}\label{sec::related}

\noindent\textbf{Probing LLMs.}
\hl{Growing concerns over the transparency and steerability of LLMs have driven recent advances in reverse-engineering LLMs by extracting interpretable features from their neural activations through probes~\citep{sharkey2025open,alain2017understanding,rogers-etal-2020-primer,hewitt-liang-2019-designing,voita-titov-2020-information,pimentel-etal-2020-information}. } 
Supervised probing typically maps neuron activations to interpretable semantics through regression or classification~\citep{gurnee2023finding, gurnee2024language, jin2024emergent, ju2024large, dong2023probing, kissane2024interpreting, templeton2024scaling, belinkov2022probing}.
For example, Gurnee \textit{et al.}~\citep{gurnee2024language} predicted the time and location of input entities from LLM activations.
Unsupervised probing, by contrast, aims to learn a dictionary of disentangled features related to more abstract concepts~\citep{engels2024not, gao2024scaling, rajamanoharan2024improving, lieberum2024gemma,mudide2024efficient,engels2024not}.
A famous example is the \textit{Golden Gate Bridge} feature identified in the Claude 3 Sonnet model~\citep{templeton2024scaling}.
While prior work focused on connecting LLM activations to external semantics, our work studies the \textit{functional topology} of neurons in LLMs, and relates this internal structure directly to language generation performance via \textit{graph probing}.

\noindent\textbf{Network Neuroscience.}
The study of functional networks in the human brain has been a central topic in neuroscience for decades~\citep{bassett2006small, bassett2017network, fotiadis2024structure, medaglia2015cognitive} which motivates this research.
Brain networks are typically constructed by correlating fMRI or EEG signals across different neural regions, and then analyzed using tools from network science~\citep{barabasi2013network}, which has revealed a range of structural and functional properties, such as small-worldness~\citep{bassett2006small}, economical wiring~\citep{bullmore2012economy}, and functional specialization~\citep{fotiadis2024structure}.
More recently, several studies have drawn parallels between LLM activations and human brain activity~\citep{toneva2019interpreting, caucheteux2023evidence, kumar2024shared, rathitopolm, mischler2024contextual, tuckute2024driving, gahot2024fmri, sun2024brain, liu2025brain}.  
For instance, Tuckute \textit{et al.}~\citep{tuckute2024driving} used GPT-2 activations to identify sentence stimuli that drive or suppress human brain responses.  
However, while these efforts focus on representational similarities, the functional \textit{topology} of neurons within LLMs and its relationship to the model’s language generation capabilities remain largely unexplored.

\section{Discussion}\label{sec::discussion}

Neurons in LLMs are connected both structurally through the model's architecture and functionally through their dynamic responses to input linguistic stimuli.
\hl{In this work, we focus on the latter and demonstrate that the language understanding and generation performance of LLMs can be reliably predicted from their functional neural topologies using graph probing, implying that LLMs develop intricate and consistent topological structures among their neurons that are fundamental to their emergent linguistic ability.}
Besides causal intervention on benchmarks validating LLMs actually leverage their internal neural topology, we also offer practical applications of neural topology in model pruning and hallucination detection.
While we have empirically shown a stable \textit{default} neural topology and hub neurons in LLMs regardless of input, we have not yet identified more nuanced structures such as motifs, or physical metrics like small-worldness and modularity within these graphs. 
It remains an open question whether such properties exist in LLMs' neural topology and play a causal role in shaping their intelligence.  
Additionally, this paper evaluates LLMs with up to 14B parameters due to computational cost, while leaving even larger models for future work.

Graph probing raises many interesting directions for future research.  
\hl{While we have linked neural topology to general linguistic ability, and discovered functional specialization and subject-specific neurons (Appendix \ref{app::functional_specialization}), it remains unclear how these specialized structures emerge during LLMs' learning process.}
Additionally, recent advances in enhancing LLMs' reasoning abilities~\citep{guo2025deepseek} raise a natural question: does reasoning alter, or is it constrained by, neural topology?  
Finally, graph probing is model-agnostic and can be extended to LLMs of distinct architectures or even models other than LLMs, for example to vision-language models (VLMs).  
We provide preliminary results of graph probing for cross-LLM matching and LLM fingerprinting, as well as probing VLMs in Appendix \ref{app::graph_matching}-\ref{app::vlm_matching} with causal intervention as well, while further efforts are required to achieve deeper insights into their multi-modal understanding and generation capabilities.
In all, we believe graph probing offers a promising lens for understanding AI models and ultimately guiding their improvement in an reliable and safe way.

\section*{Impact Statement}
This paper presents work whose goal is to advance the field of Machine Learning by improving our understanding of the internal mechanisms of large language models.
Our work relies exclusively on publicly available, pre-trained models (e.g., GPT-2, Pythia, Qwen2.5) and standard academic benchmarks (e.g., MMLU, TruthfulQA, OpenWebText) to ensure reproducibility and avoid the use of sensitive or private data.
We believe our findings have several positive societal implications. The methods for hallucination detection directly contribute to the development of more reliable, truthful, and safe AI systems. Similarly, our work on model pruning promotes computational efficiency, which can reduce the environmental impact of AI and make powerful models more accessible.

\bibliography{icml2026_conference}
\bibliographystyle{icml2026}

\clearpage
\newpage
\appendix
\section{Appendix}

\subsection{Graph Probing Configuration}\label{app::method}

\noindent\textbf{Hyperparameters.}  
We train graph probes using the Adam optimizer~\citep{adam} with mean squared error (MSE) loss, as defined in Equation~(\ref{eq::mse}).  
The initial learning rate is set to 0.00001, with a batch size of 16.  
We set the hidden dimension for MLP probes $d$ as 32.
We apply a learning rate decay strategy, reducing the rate by a factor of 0.1 if the loss does not improve for 5 consecutive epochs.  
Each model is trained for up to 100 epochs, with early stopping triggered if no improvement is observed for 20 epochs.  
Dropout is not used, as preliminary experiments showed no significant impact on regression performance.

\noindent\textbf{Computational Resources.}  
LLM inference for computing neural topologies and perplexity scores requires GPUs with large memory.  
All experiments were conducted on a Linux server equipped with 8 NVIDIA A100 GPUs (80GB memory each).  
In contrast, training graph probes is relatively lightweight and can be performed on a single GPU with 16GB memory in less than 1 hour.

\subsection{Experimented LLMs}\label{app::llm}
We run graph probing experiments on a diverse range of LLMs across three different families, with the numper of parameters ranging from 124M to 14B.
Basic information of these experimented LLMs is summarized in Table \ref{tab::llms}.

\begin{table*}[h]
\centering
\caption{Basic information of the experimented LLMs.}\label{tab::llms}
\begin{tabular}{lcccc}
\toprule
\textbf{LLM family} & \textbf{\#params} & \textbf{\#layers} & \textbf{\#neurons per layer} & \textbf{experimented layer id} \\
\midrule
\multirow{2}{*}{GPT-2} 
& 124M & 12 & 768 & 1-12\\
& 774M & 36 & 1280 & 18 \\
\hline
\multirow{3}{*}{Pythia} 
& 160M & 12 & 768 & 1-12 \\
& 1.4B & 24 & 2048 & 12 \\
& 2.8B & 32 & 2560 & 16 \\
\hline
\multirow{4}{*}{Qwen2.5}
& 0.5B & 24 & 896 & 1-12 \\
& 3B   & 36 & 2048 & 18 \\
& 7B   & 28 & 3584 & 14 \\
& 14B  & 48 & 5120 & 24 \\
\bottomrule
\end{tabular}
\end{table*}

\subsection{Datasets}\label{app::dataset}

We conduct graph probing experiments using the OpenWebText dataset~\citep{Gokaslan2019OpenWeb}.
For each dataset, we randomly sample 10,000 text sequences to construct neural connectivity graphs.
Each sample is generated by merging and tokenizing raw text until it reaches a length between 256 and 1024 tokens, which defines the length of the corresponding neural activity time series used for computing pairwise correlations.  
We then construct a text-responsive neural connectivity graph for each sample and compute its associated perplexity score.  
To remove outliers that distort the distribution, we filter out the top 1\% and bottom 1\% of samples based on perplexity.  
Finally, we normalize all perplexity values to the range $[0, 1]$ by subtracting the minimum perplexity and dividing by the observed range.
Summary statistics for the constructed datasets are provided in Table~\ref{tab::datasets}.

\begin{table*}[h]
\centering
\caption{Basic information of constructed graph probing datasets.}\label{tab::datasets}
\begin{tabular}{llccc}
\toprule
\textbf{LLM family} & \textbf{\#tokens} & \textbf{\#graphs} & \textbf{\#training graphs} & \textbf{\#test graphs}\\
\midrule
GPT-2      & 7,020,215 & 10,384 & 8,308 & 2,076\\
Pythia     & 6,798,668 & 10,441 & 8,353 & 2,088 \\
Qwen2.5    & 7,935,555 & 11,452 & 9,162 & 2,290\\
\bottomrule
\end{tabular}
\end{table*}

\subsection{Graph neural network probe}\label{app::gcn}
To reduce the number of probe parameters, we adopt a graph neural network (GNN)-based probe that encodes each node by aggregating neighborhood information through convolutional message passing on the graph~\citep{kipf2017gcn, fey2019pyg}.  
We employ the ReLU activation function~\citep{fukushima1975cognitron} between graph convolution layers and use average and maximum pooling to summarize node-level embeddings into a graph-level representation.
Given a connectivity matrix $\mathbf{A}$ induced by feeding a tokenized sequence $X$ to an LLM, where each element $a_{ij}$ denotes the functional connectivity (Pearson correlation coefficient) between neurons $i$ and $j$, our probe produces the graph representation $\mathbf{z}$ as follows:
\begin{align}
    &\mathbf{\Phi}^{0} \in \mathbb{R}^{n \times d},\label{eq::gnn1}\\
    &\mathbf{\Phi}^{l} = \mathtt{ReLU}(\mathbf{A}\mathbf{\Phi}^{l-1}\mathbf{\Theta}^l),l=1,\ldots,L,\label{eq::gnn2}\\
    & \mathbf{z} = \mathtt{AVG}\{\mathbf{\Phi}^{L}_{1,:},\ldots,\mathbf{\Phi}^{L}_{n,:}\} \parallel \mathtt{MAX}\{\mathbf{\Phi}^{L}_{1,:},\ldots,\mathbf{\Phi}^{L}_{n,:}\},\label{eq::gnn3}
\end{align}
where $\mathbf{\Phi}^{0}\in \mathbb{R}^{n \times d}$ denotes the initial learnable node embeddings, $\mathbf{\Theta}^l \in \mathbb{R}^{d \times d}$ is the weight matrix of the $l$-th layer in the GNN with $L$ total layers, and $d$ is a hidden dimensionality hyperparameter.
We then feed the graph representation $\mathbf{z}\in \mathbb{R}^{2d}$ into a multi-layer perceptron (MLP)~\citep{rumelhart1986learning} to predict the perplexity associated with the input tokenized sequence $X$:
\begin{equation}
    \hat{p} = \mathbf{W}_2 \cdot \mathtt{ReLU}(\mathbf{W}_1 \cdot \mathbf{z}^T),
\end{equation}
where $\hat{p}$ is the predicted perplexity, and $\mathbf{W_1} \in \mathbb{R}^{d\times 2d}, \mathbf{W_2} \in \mathbb{R}^{1\times d}$ are learnable weights of the MLP.

\subsection{Neural topology on texts of different subjects.}\label{app::functional_specialization}

\begin{figure*}[h]
  \centering
  \includegraphics[width=0.95\linewidth]{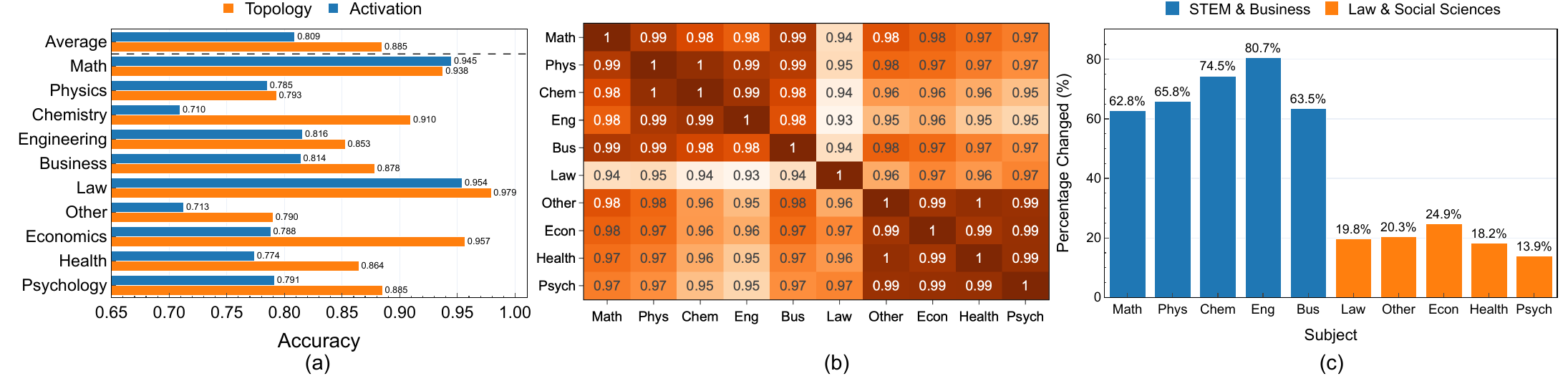}
  \vspace{-5px}
  \caption{(a) Subject classification accuracy probed from neural topology and activation of Qwen2.5-0.5B model on MMLU benchmark. (b) Correlation of neural topology of Qwen2.5-0.5B on different subjects. (c) Percentage of changed questions in each subject by intervening neuron \#894 in layer 12 of Qwen2.5-0.5B.}\label{fig::mmlu-domain}
  \vspace{-10px}
\end{figure*}

Beyond the intrinsic default network, we investigated whether LLMs form specialized networks for different knowledge domains, a plausible consequence of their multi-disciplinary pretraining. 
Such task-invoked networks have been observed in human brains~\citep{cole2014intrinsic}.
To test this, we leveraged the MMLU-Pro benchmark~\citep{wang2024mmlu} to see if a query's subject can be predicted from its corresponding neural topology. 
We framed this as a 10-way classification problem using the 10 subjects with the most samples. 
For computational efficiency, we applied a simple linear probe to the flattened adjacency matrix, optimized with a cross-entropy loss, similar to our hallucination detection setup.
The results of Qwen2.5-0.5B model, shown in Figure \ref{fig::mmlu-domain}(a), reveal that neural topology is highly indicative of the subject matter. 
The topology-based linear probe outperformed a baseline probe on neural activations by 9.39\% on average. 
This advantage held true for nearly every individual subject, with the maximum performance gap exceeding 28.16\%. 
These findings strongly suggest that LLMs develop distinct and linearly separable topological patterns for different knowledge domains, allowing for easy extraction of the context or subject being processed.

To provide intuitive evidence for subject-invoked topologies, we calculated the average neural topology for each of the 10 subjects and visualized their pairwise correlations in Figure \ref{fig::mmlu-domain}(b). 
The results are striking: the correlation matrix largely mirrors the conceptual similarities between these academic fields. 
While all pairwise correlations exceed 0.9 (reaffirming the existence of a strong default network), the variations reveal an intriguing structure. 
For example, the topologies for Math, Physics, and Engineering are highly inter-correlated (>0.98), and all share a much lower correlation (<0.95) with Law. 
More broadly, we observe a clear clustering that separates STEM and social science subjects, consistent with commonsense knowledge. 
To seek causal evidence for this phenomenon, we performed a sweeping intervention on individual neurons. 
We systematically pinned a neuron's activation to a fixed value in \{-2, -1, 0, 1, 2\} and measured the resulting change in the model's output on queries from different subjects. 
This allowed us to identify neuron \#894 in the 12th layer of Qwen2.5-0.5B, a "STEM neuron" whose activation state has a disproportionate impact on STEM-related subjects. 
As illustrated in Figure \ref{app::functional_specialization}(c), intervening on this single neuron induced 3.57 times more change in model outputs for STEM queries than for social science queries. 
While preliminary, these correlational and causal findings strongly suggest that functional specialization is an emergent property of LLMs, hinting at shared organizational principles between artificial and biological intelligence.

\subsection{Model matching and fingerprinting}\label{app::graph_matching}

\begin{figure*}[t]
  \centering
  \includegraphics[width=0.8\linewidth]{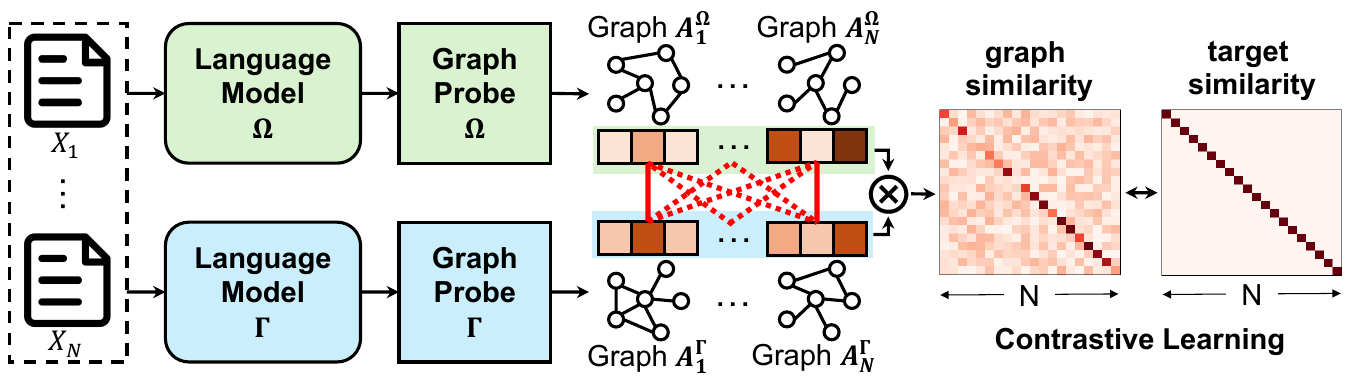}
  \vspace{-10px}
  \caption{An overview of graph matching. We learn representations of neural topologies derived from two different LLMs processing the same text dataset. We then perform contrastive learning on the graph representations such that matching pairs are more similar by inner product.}\label{fig::matching}
  \vspace{-15px}
\end{figure*}

In this section, we further investigate potential topological similarity across different LLMs to answer the following question: can we detect the genetic relationships of LLMs (\textit{e.g.} same model family, or finetuning) from their internal neural topology?
To this end, we extend graph probing with contrastive learning to perform \textit{graph matching}, as illustrated in Figure~\ref{fig::matching}.  
Specifically, suppose we feed a batch of $B$ token sequences into two LLMs, $\Omega$ and $\Gamma$.
We compute the corresponding neural connectivity graphs and use two GNN probes to encode them into representations $\mathbf{Z}_\Omega = [\mathbf{z}^\Omega_1,\ldots,\mathbf{z}^\Omega_B]$ and $\mathbf{Z}_\Gamma = [\mathbf{z}^\Gamma_1,\ldots,\mathbf{z}^\Gamma_B]$.
Matching is implemented using a contrastive cross-entropy (CE) loss that encourages alignment between graph representations from the same input texts:
\begin{align}
    &\mathcal{S} = \mathtt{MAT\_MUL}(\mathbf{Z}^T_\Omega, \mathbf{Z}_\Gamma),\quad \mathcal{T}=\mathtt{IDENTITY}(B),\\
    &\mathcal{L} = \sum_{i=1}^B{\mathtt{CE}(\mathcal{S}_{i,:}, \mathcal{T}_{i,:})} + \sum_{j=1}^B{\mathtt{CE}(\mathcal{S}_{:,j}, \mathcal{T}_{:,j})},
\end{align}
where $\mathcal{S}$ is the similarity matrix by taking inner product of graph representations and $\mathtt{IDENTITY}$ indicates identity matrix which is the target $\mathcal{T}$ for graph matching.
After training the graph probes contrastively on a shared set of training texts, the out-of-sample graph matching performance serves as an indicator of neural topology similarity between two LLMs. 
To evaluate this, we adopt the commonly used GAUC metric~\citep{li2019graph}. 
Specifically, given the predicted similarity matrix $\mathcal{S} \in \mathbb{R}^{N \times N}$ and the target similarity matrix $\mathcal{T}=\mathtt{IDENTITY}(N)$, GAUC for graph matching is calculated as follow:
\begin{align}
    &\mathtt{GAUC} = \frac{1}{2N}\sum_{i=1}^N{(\mathtt{AUC}(\mathcal{S}_{i,:}, \mathcal{T}_{i,:})+\mathtt{AUC}(\mathcal{S}_{:,i}, \mathcal{T}_{:,i}))}.
\end{align}

\begin{table}[t]
\caption{Graph matching performance (GAUC $\times$100) between different LLMs.}\label{table::matching}
\centering
\begin{tabular}{lllcc}
\toprule
\textbf{Matching} & \textbf{LLM $\mathbf{\Omega}$} & \textbf{LLM $\mathbf{\Gamma}$} & \textbf{GAUC} \\
\midrule
\multirow{3}{*}{\textbf{Self}}
 & GPT-2 & GPT-2 & 98.64 \\
 & Pythia & Pythia & 96.92 \\
 & Qwen2.5 & Qwen2.5 & 99.24 \\
\hline
\multirow{3}{*}{\textbf{Generation}}
 & Qwen2.5 & Qwen2 & 93.27 \\
 & Qwen2.5 & Qwen1.5 & 96.10 \\
 & Qwen2 & Qwen1.5 & 94.21 \\
 \hline
\multirow{3}{*}{\textbf{Family}}
 & GPT-2 & Pythia & 92.00 \\
 & GPT-2 & Qwen2.5 & 91.11 \\
 & Pythia & Qwen2.5 & 87.39 \\
\bottomrule
\end{tabular}
\end{table}

Table~\ref{table::matching} presents the graph matching results.
As a sanity check, we first perform \textit{self-matching} using the same LLM, and the results indeed show that GAUC is close to 1.0, validating the rationality of our methodology.
We then incorporate two configurations: (1) LLMs within the same family but from different generations, (2) LLMs across different families.
Given the reduced architectural and training data differences within the same model family, cross-generation LLMs are \textit{genetically} closer than cross-family LLMs.
As expected, cross-generation matching significantly outperforms cross-family matching with the average and maximum GAUC gap up to 4.84\% and 9.97\%, confirming the effectiveness of graph matching in detecting genetic closeness of LLMs.

\begin{figure}[t]
  \centering
    \includegraphics[width=\linewidth]{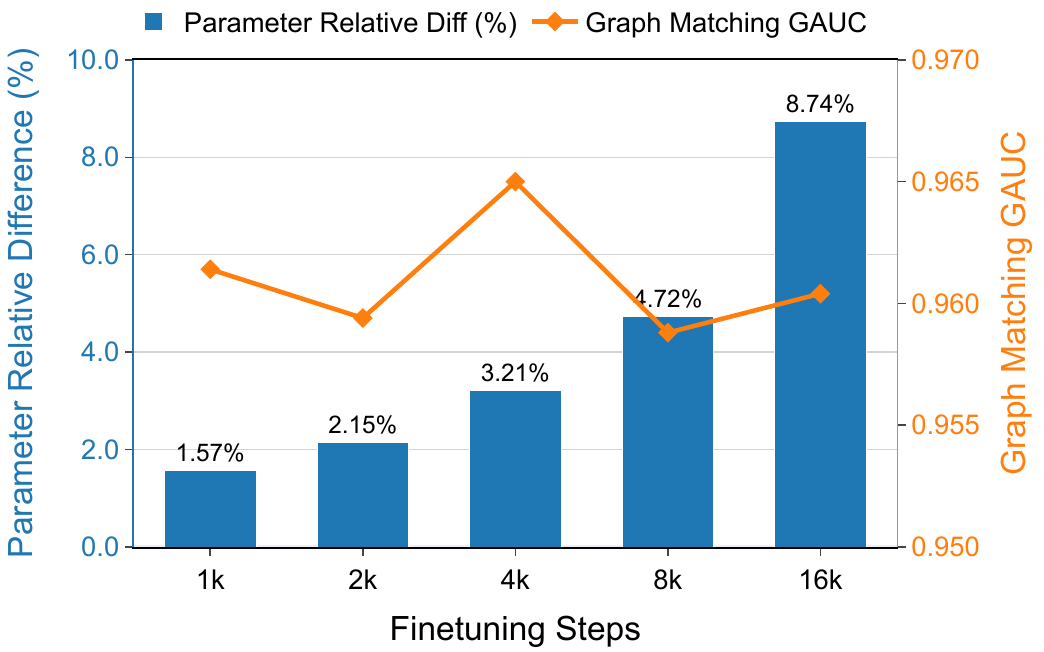}
    \caption{Relative differences of parameters (blue) and graph matching accuracy (orange) of Pythia-160M at different checkpoints.}\label{fig::fingerprint}
\end{figure}

The graph matching extension enables a direct application in LLM fingerprinting~\citep{xu2024instructional,russinovich2024hey}, which is crucial for protecting intellectual property.
To test this, we perform graph matching on the Pythia-160M model, using checkpoints at 1k, 2k, 4k, 8k, and 16k steps of continued training from a base checkpoint (127k). 
Figure \ref{fig::fingerprint} illustrates the graph matching accuracy measured by GAUC, as well as the relative differences of weight values.
Despite significant parameter drift, where the weight difference after 16k steps is 5.57 times greater than after 1k steps, the topological signature remains nearly unchanged, with our method achieving a graph matching GAUC above 0.96 in all cases. 
This suggests that neural topology can serve as a robust fingerprint, resilient to weight modifications from finetuning, which we propose as a significant avenue for future work.

\subsection{Graph probing on vision language models}\label{app::vlm_probing}

Unlike LLMs, which contain only textual hidden states, Vision-Language Models (VLMs) jointly encode image and text features within a shared transformer backbone, producing multimodal hidden states that remain underexplored. To study their internal topology, we extract the hidden representations at each layer and compute correlations as in Equations (\ref{eq::1}-\ref{eq::3}), yielding a connectivity matrix whose co-activation time series span both modalities.

\begin{figure}[h]
     \centering
     \begin{subfigure}[b]{0.45\textwidth}
         \centering
         \includegraphics[width=\textwidth]{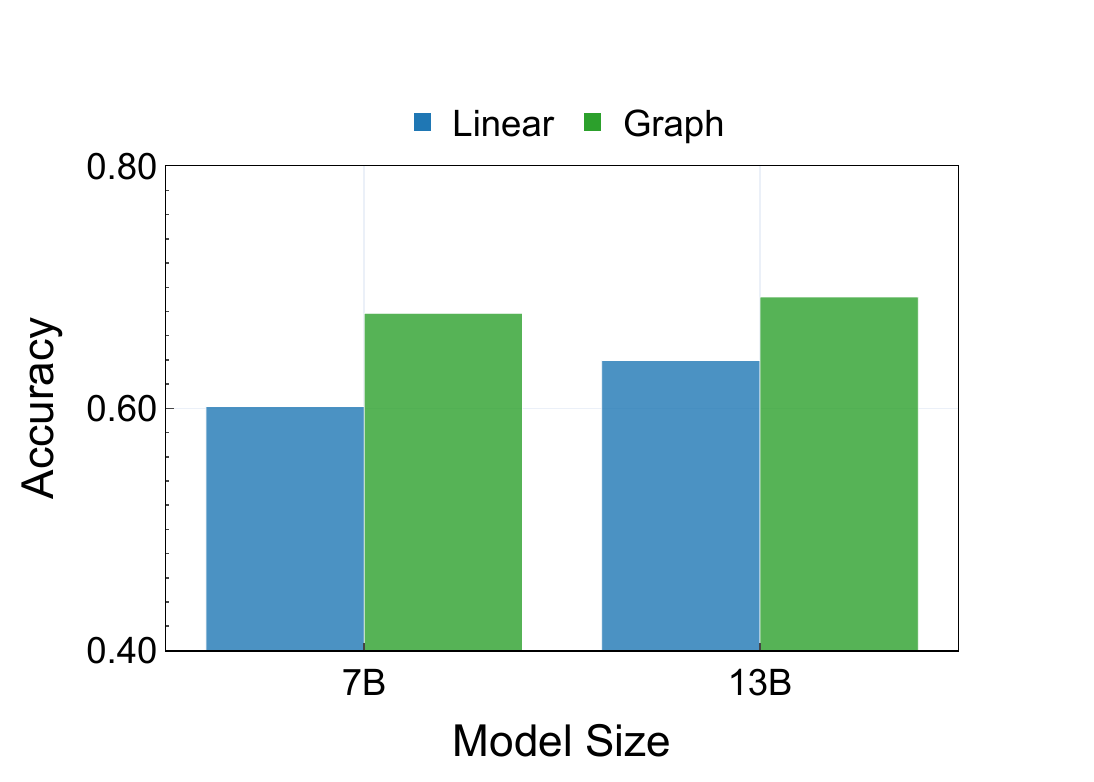}
         \vspace{-15px}
         \caption{}
         \label{fig::app-clevr}
     \end{subfigure}
     \hfill
     \begin{subfigure}[b]{0.45\textwidth}
         \centering
         \includegraphics[width=\textwidth]{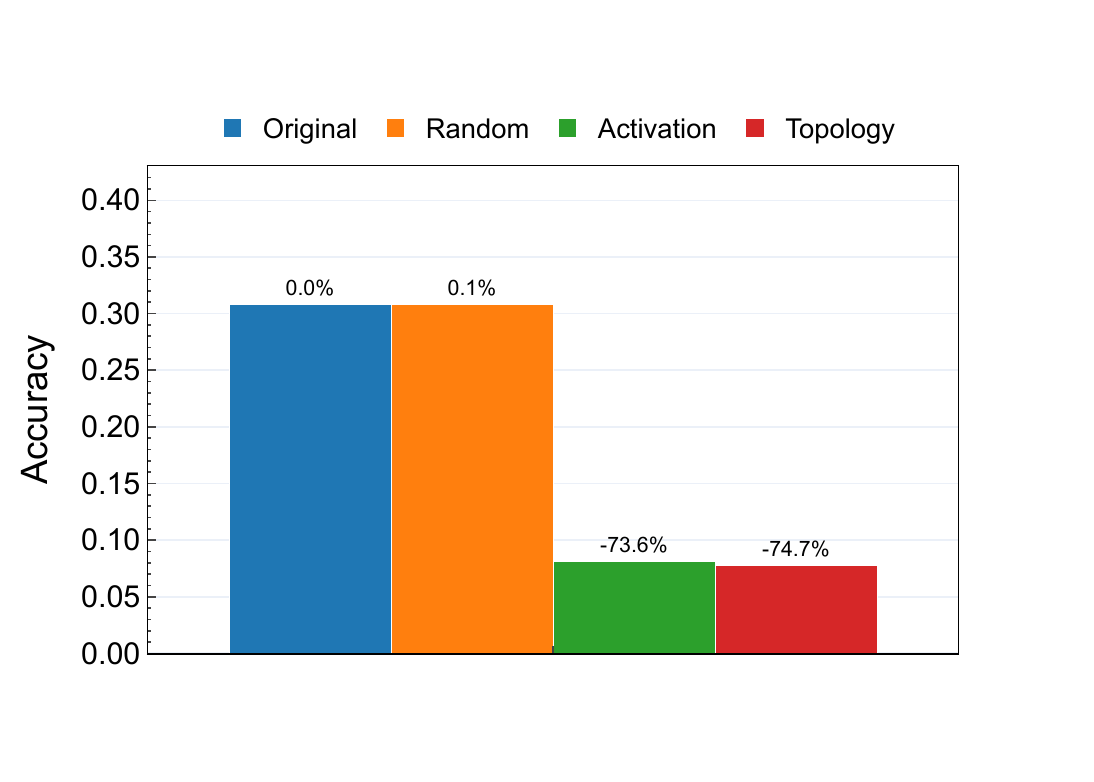}
         \vspace{-15px}
         \caption{}
         \label{fig::app-intervention}
     \end{subfigure}
     \vspace{-10px}
     \caption{(a) Out-of-sample probing performance on LLaVA-v1.5 of different sizes. (b) Accuracy on CLEVR benchmark of LLaVA-v1.5-7b under different interventions of top 1\% neurons.}\label{fig::vlm}
\end{figure}

We evaluated neural topology in VLMs by probing different sizes of LLaVA-v1.5~\citep{liu2024improved}. For this setup, we adapted the CLEVR dataset~\citep{johnson2016clevr} into an object-counting classification task of 10,000 randomly sampled images with labels corresponding to the number of objects (integers 3-10). GNN-based probes (Equations (\ref{eq::gnn1}-\ref{eq::gnn3})) were trained with cross-entropy loss on neural topology graphs constructed at a fixed density of 0.01. As shown in Figure \ref{fig::vlm}(a), graph probing consistently outperformed linear activation probing across both the 7B and 13B models, validating that neural topology is more informative than neural activation in VLMs regarding visual understanding capabilities.

To further evaluate the role of neural topology, we conducted intervention experiments on VLMs on the same dataset. Instead of restricting the probe to classification, we asked the model to generate numeric outputs and then ablated the top 1\% of nodes, selected by degree, activation, or at random, by setting their values to zero. Figure \ref{fig::vlm}(b) demonstrates that ablating top nodes by either activation or topology reduced accuracy far below random ablation and the original baseline, with topology producing the largest drop.

\begin{table}[t]
\centering
\caption{Graph matching performance (GAUC $\times 100$) for different VLMs}
\label{tab:llava-matching}
\begin{tabular}{lll c}
\toprule
\textbf{VLM $\mathbf{\Omega}$ Modality} & \textbf{VLM $\mathbf{\Gamma}$ Modality} & \textbf{GAUC} \\
\midrule
LLaVA Image+Text & LLaVA Image+Text & 95.98 \\
LLaVA Text & LLaVA Image & 81.88 \\
LLaMA & LLaVA Text  & 68.03 \\
\bottomrule
\end{tabular}
\end{table}

\subsection{Graph matching on vision language models}\label{app::vlm_matching}

We employ graph matching on VLMs to evaluate topological similarity across modalities and to test whether multimodal training reorganizes neural topology relative to unimodal language models. We use the MS-COCO dataset~\citep{lin2014mscoco}, where paired images and captions describe the same content, providing a natural basis for structural alignment. Text graphs are constructed by masking visual tokens, and image graphs by isolating patch embeddings, and multimodal graphs by combining both. We fix graph density at 0.01 and use the same contrastive loss as in the LLM matching experiments.

Table \ref{tab:llava-matching} reports graph matching scores. As a baseline, multimodal LLaVA graphs compared against themselves yield near-perfect alignment (95.98). Matching LLaVA text-only against image-only produces a notable drop (81.88), suggesting that while text and image graphs share broad structural similarities, each retains specialized organization patterns that multimodal training does not fully unify. In contrast, matching LLaMA text against LLaVA text yields a lower score (68.03), indicating that multimodal training reshapes internal topology beyond unimodal pretraining.

\end{document}